\documentclass[twoside,11pt,hidelinks]{article}

\usepackage[a4paper,%
            left=1in,right=1in,top=1.1in,bottom=1.1in]{geometry}

\usepackage[round]{natbib}
\pdfoutput=1
\usepackage{hyperref}
\hypersetup{
    colorlinks=false,
    linkcolor=black,
    citecolor=black,
    filecolor=black,
    urlcolor=black,
}

\usepackage[utf8]{inputenc}
\usepackage[T1]{fontenc}
\usepackage{booktabs}
\usepackage{array}

\usepackage{caption}
\usepackage{tabularx}
\usepackage{graphicx}
\usepackage{booktabs}
\usepackage{lscape}
\usepackage{placeins}
\usepackage{colortbl}
\usepackage{dsfont}
\usepackage{multirow}
\usepackage{listings}
\usepackage{adjustbox}
\usepackage{setspace}
\usepackage{url}

\makeatletter
\def\maxwidth{ %
  \ifdim\Gin@nat@width>\linewidth
    \linewidth
  \else
    \Gin@nat@width
  \fi
}
\makeatother

\usepackage{etoolbox}
\appto\UrlBreaks{\do\a\do\b\do\c\do\d\do\e\do\f\do\g\do\h\do\i\do\j
\do\k\do\l\do\m\do\n\do\o\do\p\do\q\do\r\do\s\do\t\do\u\do\v\do\w
\do\x\do\y\do\z}

\newcommand{\argmin}[1]{\underset{#1}{\operatorname{arg}\,\operatorname{min}}\;}

\usepackage{amsmath}
\usepackage{color}

\IfFileExists{upquote.sty}{\usepackage{upquote}}{}
\begin{document}

\title{Tunability: Importance of Hyperparameters of Machine Learning Algorithms}

\author{by Philipp Probst, Anne-Laure Boulesteix and Bernd Bischl}

\maketitle
\abstract{
Modern supervised machine learning algorithms involve hyperparameters that have to be set before running them. 
Options for setting hyperparameters are default values from the software package, manual configuration by the user or configuring them for optimal predictive performance by a tuning procedure. The goal of this paper is two-fold. Firstly, we formalize the problem of tuning from a statistical point of view, define data-based defaults and suggest general measures quantifying the tunability of hyperparameters of algorithms. Secondly, we conduct a large-scale benchmarking study based on 38 datasets from the OpenML platform and six common machine learning algorithms. We apply our measures to assess the tunability of their parameters. Our results yield default values for hyperparameters and enable users to decide whether it is worth conducting a possibly time consuming tuning strategy, to focus on the most important hyperparameters and to chose adequate hyperparameter spaces for tuning. 
}

\section{Introduction}

Machine learning (ML) algorithms such as gradient boosting, random forest and neural networks for regression and classification involve a number of {\it hyperparameters} that have to be set before running them. In contrast to direct, first-level model parameters, which are determined during training, these second-level {\it tuning parameters} often have to be carefully optimized to achieve maximal performance. A related problem exists in many other algorithmic areas, e.g., control parameters in evolutionary algorithms \citep{Eiben2011}. 

In order to select an appropriate hyperparameter {\it configuration} for a specific dataset at hand, users of ML algorithms can resort to default values of hyperparameters that are specified in implementing software packages or manually configure them, for example, based on recommendations from the literature, experience or trial-and-error. 

Alternatively, one can use hyperparameter {\it tuning strategies}, which are data-dependent, second-level optimization procedures \citep{Guyon2010}, which try to minimize the expected generalization error of the inducing algorithm over a hyperparameter search space of considered candidate configurations, usually by evaluating predictions on an independent test set, or by running a resampling scheme such as cross-validation \citep{Bischl2012}. For a recent overview of tuning strategies, see, e.g., \citet{Luo2016}. 

These search strategies range from simple grid or random search \citep{Bergstra2012} to more complex, iterative procedures such as
Bayesian optimization \citep{Hutter2011,Snoek2012,Bischl2017} or iterated F-racing \citep{Birattari2010,Lang2017}. 

In addition to selecting an efficient tuning strategy, 
the set of tunable hyperparameters and their corresponding ranges, scales and potential prior distributions for subsequent sampling have to be determined by the user. Some hyperparameters might be safely set to default values, if they work well across many different scenarios. 
Wrong decisions in these areas can inhibit either the quality of the resulting model or at the very least the efficiency and fast convergence of the tuning procedure.
This creates a burden for: 
\begin{enumerate}
\item ML users -- Which hyperparameters should be tuned and in which ranges?
\item Designers of ML algorithms -- How do I define robust defaults?
\end{enumerate}
We argue that many users, especially if they do not have years of practical experience in the field, here often rely on heuristics or spurious knowledge. It should also be noted that designers of fully automated tuning frameworks face at least very similar problems. 
It is not clear how these questions should be addressed in a data-dependent, automated, optimal and objective manner. In other words, the scientific community not only misses answers to these questions for many algorithms but also a systematic framework, methods and criteria, which are required to answer these questions.

With the present paper we aim at filling this gap and formalize the problem of parameter tuning from a statistical point of view, in order to simplify the tuning process for less experienced users and to optimize decision making for more advanced processes.  

After presenting related literature in section~\ref{sec:literature}, 
we define theoretical measures for assessing the impact of tuning in section~\ref{sec:methods}. 
For this purpose we (i)~define the concept of {\it default hyperparameters}, (ii)~suggest measures for quantifiying the tunability 
of the whole algorithm and specific hyperparameters based on the differences between the performance of default hyperparameters and the performance 
of the hyperparameters when this hyperparameter is set to an optimal value. Then we (iii)~address the tunability of hyperparameter combinations and joint gains, (iv)~provide theoretical definitions for an appropriate hyperparameter space on which tuning 
should be executed and (v)~propose procedures to estimate these quantities based on the results of a benchmark study  with 
random hyperparameter configurations with the help of surrogate models. In sections~\ref{sec:exp_setup} and \ref{sec:results} we illustrate these 
concepts and methods through an application. For this purpose we use benchmark results of six machine learning algorithms 
with different hyperparameters which were evaluated on 38 datasets from the OpenML platform. Finally, in the last section~\ref{sec:conclusion} 
we conclude and discuss the results. 

\section{Related literature}
\label{sec:literature}

To the best of our knowledge, only a limited amount of articles address the problem of tunability and generation of tuning search spaces. 
\citet{Bergstra2012} compute the relevance of the hyperparameters of neural networks and conclude that some 
are important on all datasets, while others are only important on some datasets. Their conclusion is primarily visual and used as an argument for why random search works better than grid search when tuning neural networks.

A specific study for decision trees was conducted by \citet{Mantovani2016} who apply standard tuning techniques to decision trees on 102 datasets and calculate differences of accuracy between the tuned algorithm and the algorithm with default hyperparameter settings. 

A different approach is proposed by \citet{Hutter2013}, which aims at identifying the most important hyperparameters via forward selection. 
In the same vein, \citet{Fawcett2016} present an \textit{ablation analysis} technique, which aims at identifying the hyperparameters that contribute the most to improved performance after tuning. 
For each of the considered hyperparameters, they compute the performance gain that can be achieved by changing its value from the initial value to the value specified in the target configuration which was determined by the tuning strategy. This procedure is iterated in a greedy forward search. 

A more general framework for measuring the importance of single hyperparameters is presented by \citet{Hutter2014}. 
After having used a tuning strategy such as sequential model-based optimization, a functional ANOVA approach is used for measuring the importance of hyperparameters. 

These works concentrate on the importance of hyperparameters on single datasets, mainly to retrospectively explain what happened during an already concluded tuning process. 
Our main focus is the generalization across multiple datasets in order to facilitate better general understanding of hyperparameter effects and better decision making for future experiments. 
In a recent paper \citet{Rijn2017} pose very similar questions to ours to assess the importance of hyperparameters across datasets. We compare  it to our approach  in section \ref{sec:conclusion}.

Our framework is based on using surrogate models, also sometimes called empirical performance models, which allow estimating the performance of arbitrary hyperparameter configurations based on a limited number of prior experiments. 
The idea of surrogate models is far from new, as it constitutes the central idea of Bayesian optimization for hyperparameter search but is also used, for example, in \citet{Biedenkapp2017} for increasing the speed of an ablation analysis and by \citet{Eggensperger2018} for speeding up the benchmarking of tuning strategies. 

\section{Methods for Estimation of Defaults, Tunability and Ranges}
\label{sec:methods}

\subsection{General notation}

Consider a target variable $Y$, a feature vector $X$, and an unknown joint distribution $P$ on $(X,Y)$, from which we have sampled a dataset $\mathcal{T}$ of $n$ observations. 
A machine learning (ML) algorithm now learns the functional relationship between $X$ and $Y$ by producing a prediction model $\hat{f}(X, \theta)$, controlled by the $k$-dimensional hyperparameter configuration $\theta = (\theta_1, ...,\theta_k)$ from the hyperparameter search space $\Theta = \Theta_1 \times ... \times \Theta_k$. 
In order to measure prediction performance pointwise between the true label $Y$ and its prediction $\hat{f}(X, \theta)$, we define a loss function $L(Y, \hat{f}(X, \theta))$. 
We are naturally interested in estimating the expected risk of the inducing algorithm, w.r.t. $\theta$ on new data, also sampled from $\mathcal{P}$:
$R(\theta) = E(L(Y,\hat{f}(X, \theta))| \mathcal{P}).$ 
This mapping encodes, given a certain data distribution, a certain learning algorithm and a certain performance measure, the numerical quality for any hyperparameter configuration $\theta$. Given $m$ different datasets (or data distributions) $\mathcal{P}_1,...,\mathcal{P}_m$, we arrive at $m$ hyperparameter risk mappings
\begin{equation}
R^{(j)}(\theta) := E(L(Y,\hat{f}(X, \theta))| \mathcal{P}_j), \qquad j = 1,...,m.
\end{equation}
For now, we assume all $R^{(j)}(\theta)$ to be known, and show how to estimate them in section~\ref{subsec:estimation}. 

\subsection{Optimal configuration per dataset and optimal defaults}
\label{sec:default}

We first define the best hyperparameter configuration for dataset $j$ as

\begin{equation}
\label{eq:optimal_config_per_ds}
\theta^{(j)\star} := \argmin{\theta \in \Theta} R^{(j)}(\theta).
\end{equation}

Defaults settings are supposed to work well across many different datasets and are usually provided by software packages, in an often ad hoc or heuristic manner. 
We propose to define an optimal default configuration, based on an extensive number of empirical experiments on $m$ different benchmark datasets, by 
\begin{equation}
\label{eq:default}
 \theta^\star := \argmin{\theta \in \Theta} g(R^{(1)}(\theta), ..., R^{(m)}(\theta)).
\end{equation}

Here, $g$ is a summary function that has to be specified. Selecting the mean (or median as a more robust candidate) would imply 
minimizing the average  (or median)  risk over all datasets. 

The measures $R^{(j)}(\theta)$ could potentially be scaled appropriately beforehand in order to make them more commensurable between datasets, e.g., one could scale all $R^{(j)}(\theta)$ to $[0,1]$ by substracting the result of a very simple baseline like a featureless dummy predictor and dividing this difference by the absolute difference between the risk of the best possible result (as an approximation of the Bayes error) and the result of the very simple baseline predictor. Or one could produce a statistical z-score by subtracting the mean and dividing by the standard deviation from all experimental results on the same dataset \citep{Feurer2018}.

The appropriateness of the scaling highly depends on the performance measure that is used. 
One could, for example, argue that the AUC does not have to be scaled as an improvement from 0.5 to 0.6 can possibly be seen as important as an improvement from 0.8 to 0.9. 
On the other hand, averaging the mean squared error on several datasets does not make a lot of sense, as the scale of the outcome of different regression problems can be very different. 
Then scaling or using another measure such as R$^2$ seems essential. 

\subsection{Measuring overall tunability of a ML algorithm}
\label{sec:tunability}

A general measure of the tunability of an algorithm per dataset can then be computed based on the difference between the risk of an overall reference configuration (e.g., either the software defaults or definition~\eqref{eq:default}) and the risk of the best possible configuration on that dataset:

\begin{equation}
\label{eq:tunability_algo}
d^{(j)} := R^{(j)}(\theta^{\star}) - R^{(j)}(\theta^{(j)\star}) , \mbox{ for } j = 1,...,m.
\end{equation}

For each algorithm, this gives rise to an empirical distribution of performance differences over datasets, which might be directly visualized or summarized to an aggregated tunability measure $d$ by using mean, median or quantiles. 

\subsection{Measuring tunability of a specific hyperparameter} 
\label{sec:tunability_par}
The best hyperparameter value for one parameter $i$ on dataset $j$, when all other parameters are set to defaults from $\theta^{\star} := (\theta_1^{\star}, ...,\theta_k^{\star})$, is denoted by

\begin{equation}
\theta^{(j)\star}_{i} := \argmin{\theta \in \Theta, \theta_{l} = \theta_l^{\star} \forall l \neq i} R^{(j)}(\theta).
\end{equation}

A natural measure for tunability of the $i$-th parameter on dataset $j$ is then the difference in risk between the above and our default reference configuration:
\begin{equation}
\label{eq:tunability_par}
d_i^{(j)} := R^{(j)}(\theta^{\star}) - R^{(j)}(\theta^{(j)\star}_{i}), \mbox{ for } j = 1,...,m, i = 1,...,k.
\end{equation}

Furthermore, we define $d_i^{(j),\text{rel}} = \frac{d_i^{(j)}}{d^{(j)}}$ as the fraction of performance gain, when we only tune $i$ compared to tuning the complete algorithm, on dataset $j$.
Again, one can calculate the mean, the median or quantiles of these two differences over the $n$ datasets, to get a notion of the overall tunability $d_i$ of this parameter. 

\subsection{Tunability of hyperparamater combinations and joint gains} 
\label{sec:tunability_inter}
Let us now consider two hyperparameters indexed as $i_1$ and $i_2$. To measure the tunability with respect to these two parameters, we define
\begin{equation}
\theta^{(j)\star}_{i_1,i_2} := \argmin{\theta \in \Theta, \theta_{l} = \theta_l^{\star} \forall l \not\in \{i_1, i_2\}} R^{(j)}(\theta),
\end{equation}
i.e., the $\theta$-vector containing the default values for all hyperparameters other than  $i_1$ and $i_2$, and the optimal combination of values for the $i_1$-th and $i_2$-th components of $\theta$.

Analogously to the previous section, we can now define the tunability of the set $(i_1, i_2)$ as the gain over the reference default on dataset $j$ as:
\begin{align}
d^{(j)}_{i_1, i_2} := &  R^{(j)}(\theta^*) - R^{(j)}(\theta^{(j)\star}_{i_1,i_2}).
\end{align}

The joint gain which can be expected when tuning not only one of the two hyperparameters individually, but both of them jointly, on a dataset $j$, can be expressed by:
\begin{align}
g^{(j)}_{i_1, i_2} := & \min\{(R^{(j)}(\theta^{(j)\star}_{i_1})), (R^{(j)}(\theta^{(j)\star}_{i_2}))\} - R^{(j)}(\theta^{(j)\star}_{i_1,i_2}).
\end{align}

Furthermore, one could be interested in whether this joint gain could simply be reached by tuning both parameters $i_1$ and $i_2$ in a univariate fashion sequentially, either in the order $i_1 \rightarrow i_2$ or $i_2 \rightarrow i_1$, and what order would be preferable.
For this purpose one could compare the risk of the hyperparameter value that results when tuning them together $R^{(j)}(\theta^{(j)\star}_{i_1,i_2})$ with the risks of the hyperparameter values that are obtained when tuning them sequentially, that means $R^{(j)}(\theta^{(j)\star}_{i_1 \rightarrow i_2})$ or $R^{(j)}(\theta^{(j)\star}_{i_2 \rightarrow i_1})$, which is done for example in \citet{Waldron2011}.

Again, all these measures should be summarized across datasets, resulting in $d_{i_1, i_2}$ and $g_{i_1, i_2}$. Of course, these approaches can be further generalized by considering combinations of more than two parameters. 

\subsection{Optimal hyperparameter ranges for tuning}
\label{sec:hypspace}
A reasonable hyperparameter space $\Theta^{\star}$ for tuning should include the optimal configuration $\theta^{(j)\star}$ for dataset $j$ with high probability. 
We denote the $p$-quantile of the distribution of one parameter regarding the best hyperparameters on each dataset $(\theta^{(1)\star})_i, ..., (\theta^{(m)\star})_i$ as $q_{i,p}$. 
The hyperparameter tuning space can then be defined as:

\begin{equation}
\label{eq:hypspace}
\Theta^{\star} := \left\{\theta \in \Theta | \forall i \in \{1,...,k\}: \theta_i \geq q_{i, p_1} \land \theta_i \leq q_{i, p_2} \right\},
\end{equation}
with $p_1$ and $p_2$ being quantiles which can be set for example to the 5~\% quantile and the 95~\% quantile. 
This avoids focusing too much on outlier datasets and 
makes the definition of the space independent from the number of datasets.

The definition above is only valid for numerical hyperparameters. In case of categorical variables one could use similar rules, for example 
only including hyperparameter values that were at least once or in at least 10~\% of the 
datasets the best possible hyperparameter setting. 

\subsection{Practical estimation}
\label{subsec:estimation}

In order to practically apply the previously defined concepts, two remaining issues need to be addressed: a) We need to discuss how to obtain $R^{(j)}(\theta)$; and b) in \eqref{eq:optimal_config_per_ds} and \eqref{eq:default} a multivariate optimization problem (the minimization) needs to be solved\footnote{All other previous optimization problems are univariate or two-dimensional and can simply be addressed by a simple technique like a fine grid search }. 

For a) we estimate $R^{(j)}(\theta)$ by using surrogate models $\hat{R}^{(j)}(\theta)$, and replace the original quantity by its estimator in all previous formulas. Surrogate models for each dataset $j$ are based on a meta dataset. This is created by evaluating a large number of configurations of the respective ML method. The surrogate regression model then learns to map a hyperparameter configuration to estimated performance. For b) we  solve the optimization problem -- now cheap to evaluate, because of the surrogate models -- through black-box optimization.

\section{Experimental setup}
\label{sec:exp_setup}

In this section we give an overview about the experimental setup that is used for obtaining  surrogate models, tunability measures and tuning spaces.

\subsection{Datasets from the OpenML platform}
\label{sec:datasets}
Recently, the OpenML project \citep{OpenML2013} has been created as a flexible online platform that allows ML scientists to share their data, corresponding tasks and results of different ML algorithms.  
We use a specific subset of carefully curated classification datasets from the OpenML 
platform called \textit{OpenML100} \citep{Bischl20172}.
For our study we only use the 38 binary classification tasks that do not contain any missing values. 

\subsection{ML Algorithms}
\label{sec:algorithms}
The algorithms considered in this paper are common methods for supervised learning. 
We examine elastic net (\texttt{glmnet}), decision tree (\texttt{rpart}), k-nearest neighbors (\texttt{kknn}), 
support vector machine (\texttt{svm}), random forest (\texttt{ranger}) and gradient boosting (\texttt{xgboost}). For more details about the used software packages see \citet{Kuhn2018}. 
An overview of their considered hyperparameters is displayed in Table \ref{tab:parameter}, including respective data types, box-constraints and a potential transformation function.

\begin{table}
\centering
\begin{tabular}{llrrrr}
  \hline
Algorithm & Hyperparameter & Type & Lower & Upper & Trafo \\
  \hline
\rowcolor[gray]{0.9} \hline glmnet &  &  &  &  &   \\ 
   & alpha & numeric & 0 & 1 & - \\ 
   & lambda & numeric & -10 & 10 & $2^x$ \\ 
  \rowcolor[gray]{0.9}rpart &  &  &  &  &  \\ 
   & cp & numeric & 0 & 1 & - \\ 
   & maxdepth & integer & 1 & 30 & - \\ 
   & minbucket & integer & 1 & 60 & - \\ 
   & minsplit & integer & 1 & 60 & - \\ 
  \rowcolor[gray]{0.9}kknn & - & - &  &  &  \\ 
   & k & integer & 1 & 30 & - \\ 
  \rowcolor[gray]{0.9}svm &  &  &  &  &  \\ 
   & kernel & discrete & - & - & - \\ 
   & cost & numeric & -10 & 10 & $2^x$ \\ 
   & gamma & numeric & -10 & 10 & $2^x$ \\ 
   & degree & integer & 2 & 5 & - \\ 
  \rowcolor[gray]{0.9}ranger &  &  &  &  &  \\ 
   & num.trees & integer & 1 & 2000 & - \\ 
   & replace & logical & - & - & - \\ 
   & sample.fraction & numeric & 0.1 & 1 & - \\ 
   & mtry & numeric & 0 & 1 & $x \cdot p$ \\ 
   & respect.unordered.factors & logical & - & - & - \\ 
   & min.node.size & numeric & 0 & 1 & $n^x$ \\ 
  \rowcolor[gray]{0.9}xgboost &  &  &  &  &  \\ 
   & nrounds & integer & 1 & 5000 & - \\ 
   & eta & numeric & -10 & 0 & $2^x$ \\ 
   & subsample & numeric & 0.1 & 1 & - \\ 
   & booster & discrete & - & - & - \\ 
   & max\_depth & integer & 1 & 15 & - \\ 
   & min\_child\_weight & numeric & 0 & 7 & $2^x$ \\ 
   & colsample\_bytree & numeric & 0 & 1 & - \\ 
   & colsample\_bylevel & numeric & 0 & 1 & - \\ 
   & lambda & numeric & -10 & 10 & $2^x$ \\ 
   & alpha & numeric & -10 & 10 & $2^x$ \\ 
   \hline
\end{tabular}
\caption{\small Hyperparameters of the algorithms. $p$ refers to the number of variables and $n$ to the number of observations. The columns \textit{Lower} and \textit{Upper} indicate the regions from which samples of these hyperparameters are drawn. 
The transformation function in the trafo column, if any, indicates how 
the values are transformed according to this function. 
The exponential transformation is applied to obtain more candidate values in regions with smaller hyperparameters because for 
these hyperparameters the performance differences between smaller values are potentially bigger than for bigger values.
The \texttt{mtry} value in \texttt{ranger} that is drawn from $[0,1]$ is transformed for 
each dataset separately. After having chosen the dataset, the value is multiplied by the number of variables and afterwards rounded up.
Similarly, for the \texttt{min.node.size} the value $x$  is transformed by 
the formula $[n^x]$ with $n$ being the number of observations of the dataset, to obtain a 
positive integer values with higher probability for smaller values (the value is finally rounded to obtain integer values).} 
\label{tab:parameter}
\end{table}

In the case of \texttt{xgboost}, the underlying package only supports numerical features, so we opted for a dummy feature encoding for categorical features, which is performed internally by the underlying packages for \texttt{svm} and \texttt{glmnet}.

Some hyperparameters of the algorithms are dependent on others. We take into account these dependencies and, for example, only 
sample a value for \texttt{gamma} for the support vector machine if the radial kernel was sampled beforehand.

\subsection{Performance estimation}
\label{sec:measures}

Several measures are regarded throughout this paper, either for evaluating our considered classification models that should be tuned, or for evaluating our surrogate regression models. As no optimal measure exists, we will compare several of them.
In the classification case, we consider AUC, accuracy and brier score.
In the case of surrogate regression, we consider R$^2$, which is directly proportional to the regular mean squared error but scaled to [0,1] and explains the gain over a constant model estimating the overall mean of all data points. 
We also compute Kendall's tau as a ranking based measure for regression. 

The performance estimation for the different hyperparameter experiments is computed through 10-fold cross-validation. For the comparison of surrogate models 10 times repeated 10-fold cross-validation is used.

\subsection{Random Bot sampling strategy for meta data}
\label{sec:search_strategy}

To reliably estimate our surrogate models we need enough evaluated configurations per classifier and data set.
We sample these points from independent uniform distributions where the respective support for each parameter is displayed in Table~\ref{tab:parameter}. Here, \textit{uniform} refers to the untransformed scale, so we sample uniformly from the interval [\textit{Lower}, \textit{Upper}] of Table~\ref{tab:parameter}.

In order to properly facilitate the automatic computation of a large database of hyperparameter experiments, we implemented a so called 
OpenML bot. 
In an embarrassingly parallel manner it chooses in each iteration a 
random dataset, a random classification algorithm, samples a random configuration and evaluates it via cross-validation. 
A subset of 500000 experiments for each algorithm and all datasets are used for our analysis here.\footnote{30 for each dataset for \texttt{kknn}}
More technical details regarding the random bot, its setup and results  can be obtained in \citet{Kuhn2018}, furthermore, for simple and permanent access the results of the bot are stored in a figshare repository \citep{Kuehn2018}. 

\subsection{Optimizing surrogates to obtain optimal defaults}
Random search is also used for our black-box optimization problems in section \ref{subsec:estimation}. For the estimation of the defaults for each algorithm 
we randomly sample 100000 points in the hyperparameter space as defined in Table~\ref{tab:parameter} and determine the configuration with the minimal average risk. 
The same strategy with 100000 random points is used to obtain the best hyperparameter setting on each dataset that is needed for the estimation of the tunability of an algorithm. 
For the estimation of the tunability of single hyperparameters we also use 100000 random points for each parameter, while for the tunability of combination of hyperparameters we only use 10000 random points to reduce runtime as this should be enough to cover 2-dimensional hyperparameter spaces. 

Of course one has to be careful with overfitting here, as our new defaults are chosen with the help of the same datasets that are used to determine the performance. 
Therefore, we also evaluate our approach via a ``10-fold cross-validation across datasets''. 
Here, we repeatedly calculate the optimal defaults based on 90\% ``training datasets'' and evaluate the package defaults and our optimal defaults -- the latter induced from the training data sets -- on the surrogate models of the remaining 10\% ``test datasets'', and compare their difference in performance.

\subsection{The problem of hyperparameter dependency}

Some parameters are dependent on other superordinate hyperparameters and are only relevant if the parameter value of this superordinate parameter was set to a specific value. 
For example \texttt{gamma} in \texttt{svm} only makes sense if the \texttt{kernel} was set to ``radial`` or \texttt{degree} only makes sense if the kernel was set to ``polynomial``. 
Some of these subordinate parameters might be invalid/inactive in the reference default  configuration, rendering it impossible to univariately tune them in order to compute their tunability score.
In such a case we set the superordinate parameter to a value which makes the subordinate parameter active, compute the optimal defaults for the rest of the parameters and compute the tunability score for the subordinate parameter with these defaults.

\subsection{Software details}

All our experiments are executed in R and are run through a combination of custom code from our random bot \citet{Kuhn2018}, the \texttt{OpenML} R package \citep{Casalicchio2017}, \texttt{mlr} \citep{Bischl2016} and \texttt{batchtools} \citep{Lang2017} for parallelization. All results are uploaded to the OpenML platform and there publicly available for further analysis. 
\texttt{mlr} is also used to compare and fit all surrogate regression models. 
The fully reproducible R code for all computations and analyses of our paper can be found on the github page: \url{https://github.com/PhilippPro/tunability}.
We also provide an interactive shiny app under \url{https://philipppro.shinyapps.io/tunability/}, which displays all results of the following section in a potentially more convenient, interactive fashion and which can simply be accessed through a web browser.

\FloatBarrier

\section{Results and discussion}
\label{sec:results}

We calculate all results for AUC, accuracy and brier score but mainly discuss AUC results here. Tables and  figures  for the other measures can be accessed in the  Appendix  and in our interactive shiny application.

\subsection{Surrogate models}
\label{sec:surrogate2}

\begin{figure}
\includegraphics[width=\maxwidth]{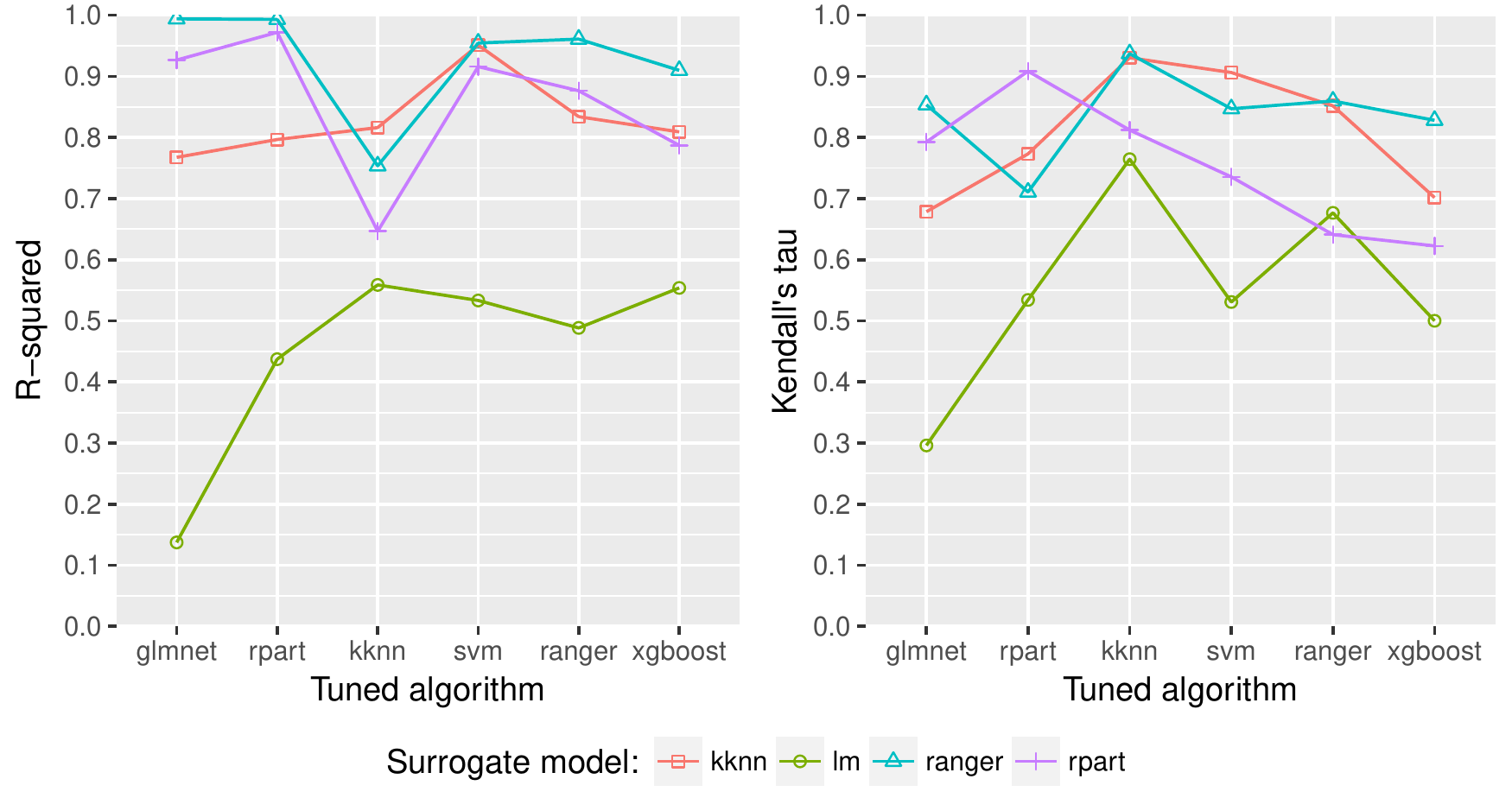} \caption{Average performances over the datasets of different surrogate models (target: AUC) for different algorithms (that were presented in \ref{sec:algorithms}).}\label{fig:surr_models}
\end{figure}

We compare different possible regression models as candidates for our surrogate models: 
the linear model (\texttt{lm}), a simple decision tree (\texttt{rpart}), \mbox{k nearest-neighbors} (\texttt{kknn}) and random forest (\texttt{ranger})\footnote{We also tried \texttt{cubist} \citep{Kuhn2016}, which provided good results but the algorithm had some technical problems for some combinations of datasets and algorithms. We did not include gaussian process which is one of the standard algorithms for surrogate models as it cannot handle categorical variables.} 
All algorithms are run with their default settings. 
We calculate 10 times repeated 10-fold cross-validated regression performance measures R$^2$ and Kendall's tau per dataset, and average these across all datasets\footnote{In case of \texttt{kknn} four datasets did not provide results for one of the surrogate models and were not used.}.
Results for AUC are displayed in Figure \ref{fig:surr_models}. 
A good overall performance is achieved by \texttt{ranger} with qualitatively similar results for other classification performance measures (see  Appendix). 
In the following we use random forest as surrogate model because it performs reasonably well and is already an established algorithm for surrogate models in the literature \citep{Eggensperger2014, Hutter2013}.

\subsubsection{Optimal defaults and tunability}

Table~\ref{tab:tunability_overall} displays our mean tunability results for the algorithms as defined in formula~\eqref{eq:tunability_algo} w.r.t. package defaults (\texttt{Def.P} column) and our optimal defaults (\texttt{Def.O}).
It also displays the improvement per algorithm when moving from package defaults to optimal defaults (\texttt{Improv}), which was positive overall. 
This also holds for \texttt{svm} and \texttt{ranger} although the package defaults are data dependent, which we currently cannot model 
(\texttt{gamma} = $1/p$ for \texttt{svm} and \texttt{mtry} = $\sqrt{p}$ for \texttt{ranger}). 
From now on, when discussing tunability, we will only do this w.r.t. our optimal defaults.

\begin{figure}[!t]
\includegraphics[width=\maxwidth]{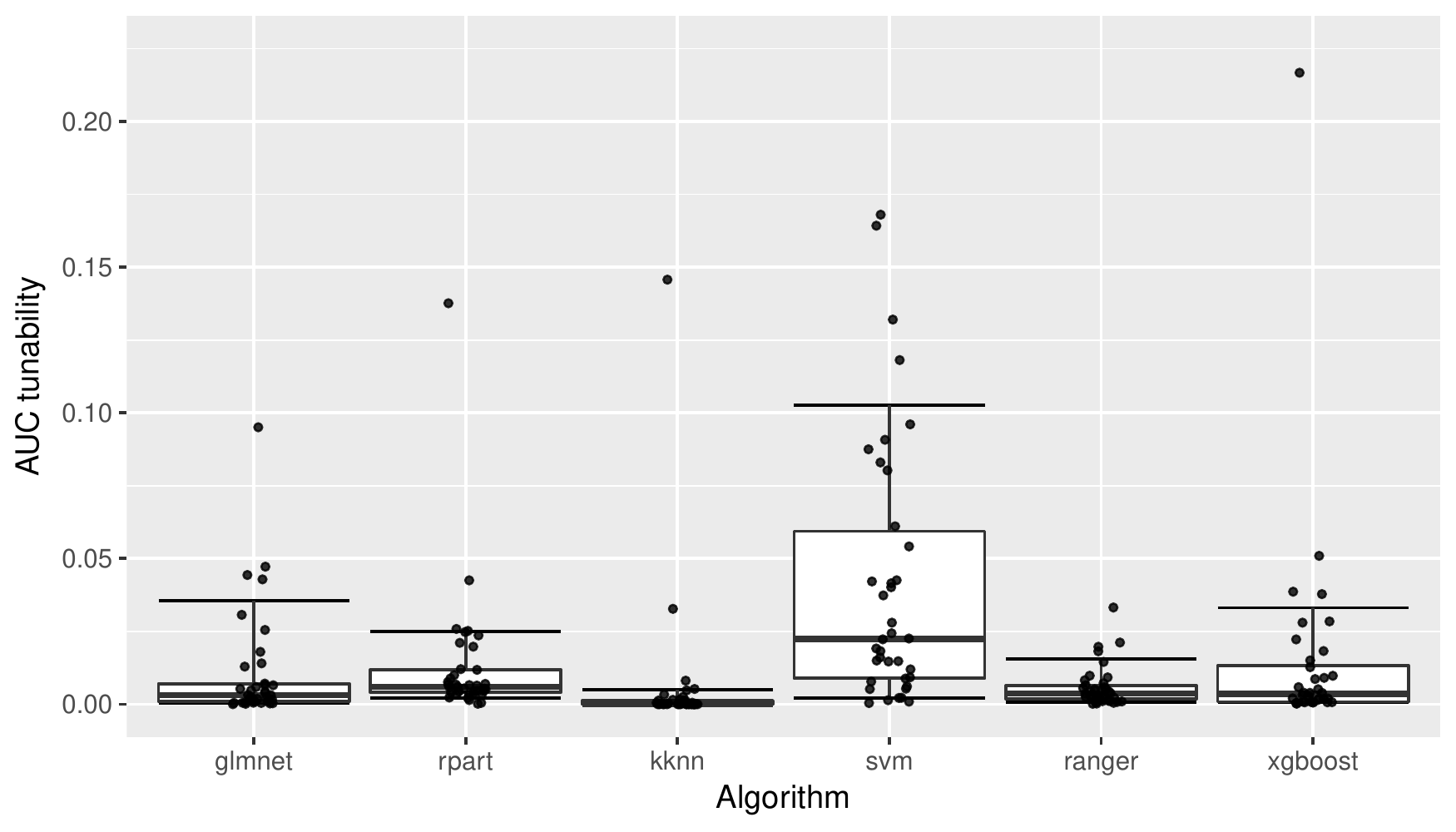} \caption[Boxplots of the tunabilities (AUC) of the different algorithms]{Boxplots of the tunabilities (AUC) of the different algorithms. The upper and lower whiskers (upper and lower line of the boxplot rectangle) are in our case defined as the 0.1 and 0.9 quantiles of the tunability scores. The 0.9 quantile indicates how much performance improvement can be expected on at least 10\% of datasets. One outlier of glmnet (value 0.5) is not shown.}\label{fig:tunability_algos}
\end{figure}

Clearly, some algorithms such as \texttt{glmnet} and \texttt{svm} are much more tunable than the others, while \texttt{ranger} is the algorithm with the smallest tunability, which is in line with common knowledge in the web community. 
In Figure \ref{fig:tunability_algos} modified boxplots of the tunabilities are depicted. For each ML algorithm, some outliers are visible, which indicates that tuning has a much higher impact on some specific datasets. 

\begin{table}
\centering
\begin{tabular}{rrrrrr}
  \hline
Algorithm & Tun.P & Tun.O & Tun.O-CV & Improv & Impr-CV \\ 
  \hline
glmnet & 0.069 & 0.024 & 0.037 & 0.045 & 0.032 \\ 
  rpart & 0.038 & 0.012 & 0.016 & 0.025 & 0.022 \\ 
  kknn & 0.031 & 0.006 & 0.006 & 0.025 & 0.025 \\ 
  svm & 0.056 & 0.042 & 0.048 & 0.014 & 0.008 \\ 
  ranger & 0.010 & 0.006 & 0.007 & 0.004 & 0.003 \\ 
  xgboost & 0.043 & 0.014 & 0.017 & 0.029 & 0.026 \\ 
   \hline
\end{tabular}
\caption{Overall tunability (regarding AUC) with the package defaults (Tun.P) and the new defaults (Tun.O) as reference, cross-validated tunability (Tun.O-CV), average improvement (Improv) and cross-validated average improvement (Impr-CV) obtained by using new defaults compared to old defaults. The (cross-validated) improvement can be calculated by the (rounded) difference between Tun.P and Tun.O (Tun.O-CV).} 
\label{tab:tunability_overall}
\end{table}

\subsubsection{Tunability of specific hyperparameters}

In Table \ref{tab:tunability} the mean tunability (regarding the AUC) of single hyperparameters as defined in Equation~\eqref{eq:tunability_par} 
in section~\ref{sec:tunability_par} can 
be seen. 
From here on, we will refer to tunability only with respect to optimal defaults.

\begin{table}[!htbp]
\centering
\begin{tabular}{rrrrrrr}
  \hline
Parameter & Def.P & Def.O & Tun.P & Tun.O & $q_{0.05}$ & $q_{0.95}$ \\ 
  \hline
\rowcolor[gray]{0.9}glmnet &  &  & 0.069 & 0.024 &  &  \\ 
  alpha & 1 & 0.403 & 0.038 & 0.006 & 0.009 & 0.981 \\ 
  lambda & 0 & 0.004 & 0.034 & 0.021 & 0.001 & 0.147 \\ 
  \rowcolor[gray]{0.9}rpart &  &  & 0.038 & 0.012 &  &  \\ 
  cp & 0.01 & 0 & 0.025 & 0.002 & 0 & 0.008 \\ 
  maxdepth & 30 & 21 & 0.004 & 0.002 & 12.1 & 27 \\ 
  minbucket & 7 & 12 & 0.005 & 0.006 & 3.85 & 41.6 \\ 
  minsplit & 20 & 24 & 0.004 & 0.004 & 5 & 49.15 \\ 
  \rowcolor[gray]{0.9}kknn &  &  & 0.031 & 0.006 &  &  \\ 
  k & 7 & 30 & 0.031 & 0.006 & 9.95 & 30 \\ 
  \rowcolor[gray]{0.9}svm &  &  & 0.056 & 0.042 &  &  \\ 
  kernel & radial & radial & 0.030 & 0.024 &  &  \\ 
  cost & 1 & 682.478 & 0.016 & 0.006 & 0.002 & 920.582 \\ 
  gamma & $1/p$ & 0.005 & 0.030 & 0.022 & 0.003 & 18.195 \\ 
  degree & 3 & 3 & 0.008 & 0.014 & 2 & 4 \\ 
  \rowcolor[gray]{0.9}ranger &  &  & 0.010 & 0.006 &  &  \\ 
  num.trees & 500 & 983 & 0.001 & 0.001 & 206.35 & 1740.15 \\ 
  replace & TRUE & FALSE & 0.002 & 0.001 &  &  \\ 
  sample.fraction & 1 & 0.703 & 0.004 & 0.002 & 0.323 & 0.974 \\ 
  mtry & $\sqrt{p}$ & $p\cdot0.257$ & 0.006 & 0.003 & 0.035 & 0.692 \\ 
  respect.unordered.factors & TRUE & FALSE & 0.000 & 0.000 &  &  \\ 
  min.node.size & 0 & 1 & 0.001 & 0.001 & 0.007 & 0.513 \\ 
  \rowcolor[gray]{0.9}xgboost &  &  & 0.043 & 0.014 &  &  \\ 
  nrounds & 500 & 4168 & 0.004 & 0.002 & 920.7 & 4550.95 \\ 
  eta & 0.3 & 0.018 & 0.006 & 0.005 & 0.002 & 0.355 \\ 
  subsample & 1 & 0.839 & 0.004 & 0.002 & 0.545 & 0.958 \\ 
  booster & gbtree & gbtree & 0.015 & 0.008 &  &  \\ 
  max\_depth & 6 & 13 & 0.001 & 0.001 & 5.6 & 14 \\ 
  min\_child\_weight & 1 & 2.06 & 0.008 & 0.002 & 1.295 & 6.984 \\ 
  colsample\_bytree & 1 & 0.752 & 0.006 & 0.001 & 0.419 & 0.864 \\ 
  colsample\_bylevel & 1 & 0.585 & 0.008 & 0.001 & 0.335 & 0.886 \\ 
  lambda & 1 & 0.982 & 0.003 & 0.002 & 0.008 & 29.755 \\ 
  alpha & 1 & 1.113 & 0.003 & 0.002 & 0.002 & 6.105 \\ 
   \hline
\end{tabular}
\caption{Defaults (package defaults (Def.P) and optimal defaults (Def.O)), 
    tunability of the hyperparameters with the package defaults (Tun.P) and our optimal defaults (Tun.O) as reference and tuning space quantiles ($q_{0.05}$ and $q_{0.95}$) 
    for different parameters of the algorithms.} 
\label{tab:tunability}
\end{table}

For \texttt{glmnet} \texttt{lambda} seems to be more tunable than \texttt{alpha}. 
In \texttt{rpart} the \texttt{minbucket} and \texttt{minsplit} parameters seem to be 
the most important ones for tuning. 
\texttt{k} in the \texttt{kknn} algorithm is very tunable w.r.t. package defaults, but not regarding optimal defaults. 
In \texttt{svm} the biggest gain in performance can 
be achieved by tuning the \texttt{kernel}, \texttt{gamma} or \texttt{degree}, while the \texttt{cost} parameter does not seem 
to be very tunable. In \texttt{ranger} \texttt{mtry} is the most tunable parameter. 
For \texttt{xgboost} there are two parameters that are quite tunable: \texttt{eta} and the \texttt{booster}. \texttt{booster} specifies if 
a tree or a linear model is trained.
The cross-validated results can be seen in Table \ref{tab:tuncv} in the  Appendix, they are quite similar to the 
non cross-validated results and for all parameters slightly higher. 

\begin{figure}[!htbp]
\includegraphics[width=\maxwidth]{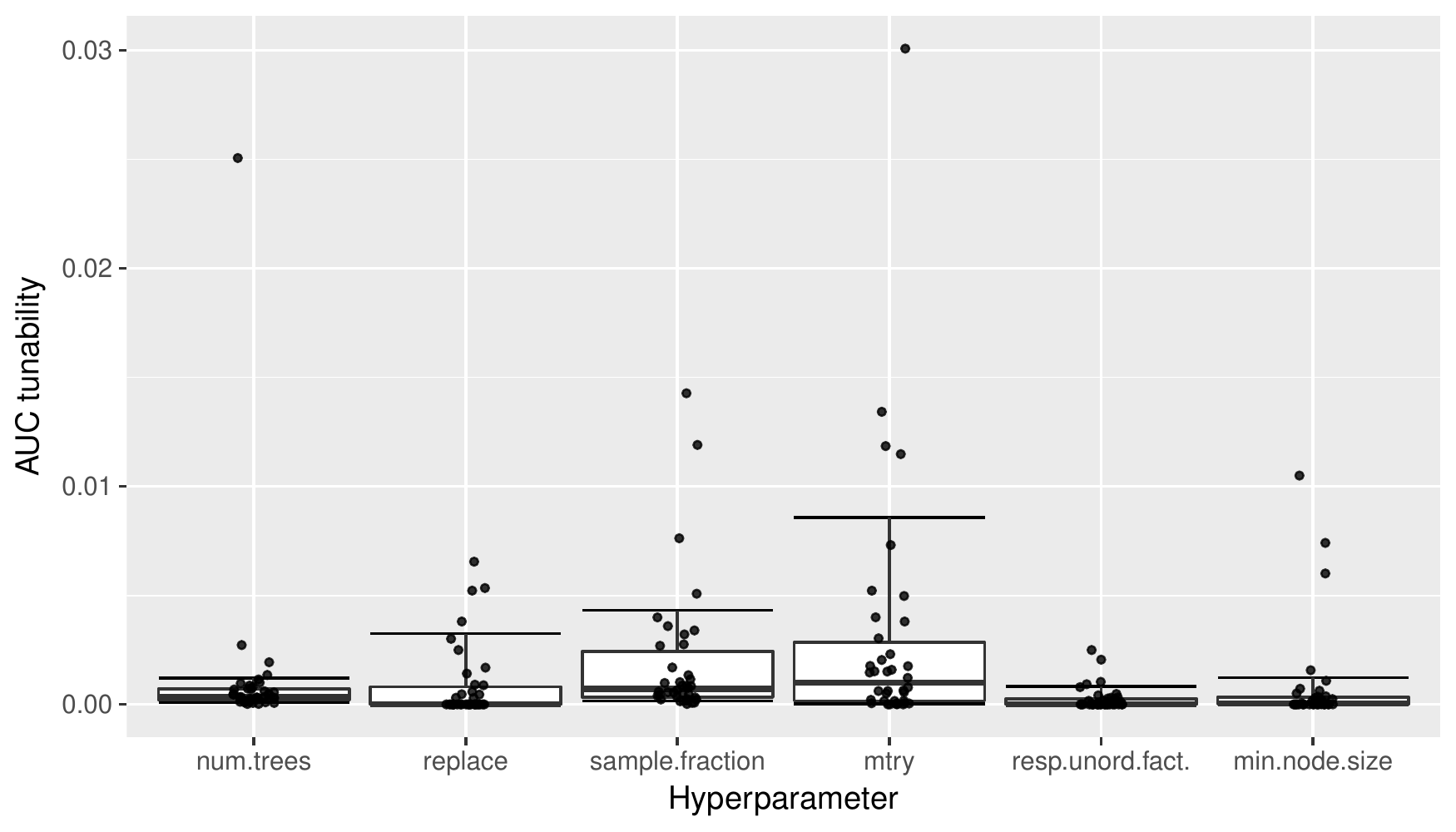} \caption[Boxplots of the tunabilities of the different parameters of \texttt{ranger}]{Boxplots of the tunabilities of the different parameters of \texttt{ranger}. Same definition of whiskers as in Figure~ \ref{fig:tunability_algos}.}\label{fig:tunability_plots}
\end{figure}

Instead of looking only at the average, as in Table \ref{tab:tunability}, one could also be interested in the distribution of the tunability of each dataset. 
As an example, Figure \ref{fig:tunability_plots} shows the tunability of each parameter of \texttt{ranger} in a boxplot. This gives a more in-depth insight 
into the tunability, makes it possible to detect outliers and to examine the skewness. 

\subsubsection{Hyperparameter space for tuning}

The hyperparameter space for tuning, as defined in Equation~\eqref{eq:hypspace} in section~\ref{sec:hypspace} and based on the 0.05 and 0.95 quantiles, is displayed in Table~\ref{tab:tunability}.
All optimal defaults are contained in this hyperparameter space while some of the package defaults are not. 

\begin{figure}[!htbp]
\includegraphics[width=\maxwidth]{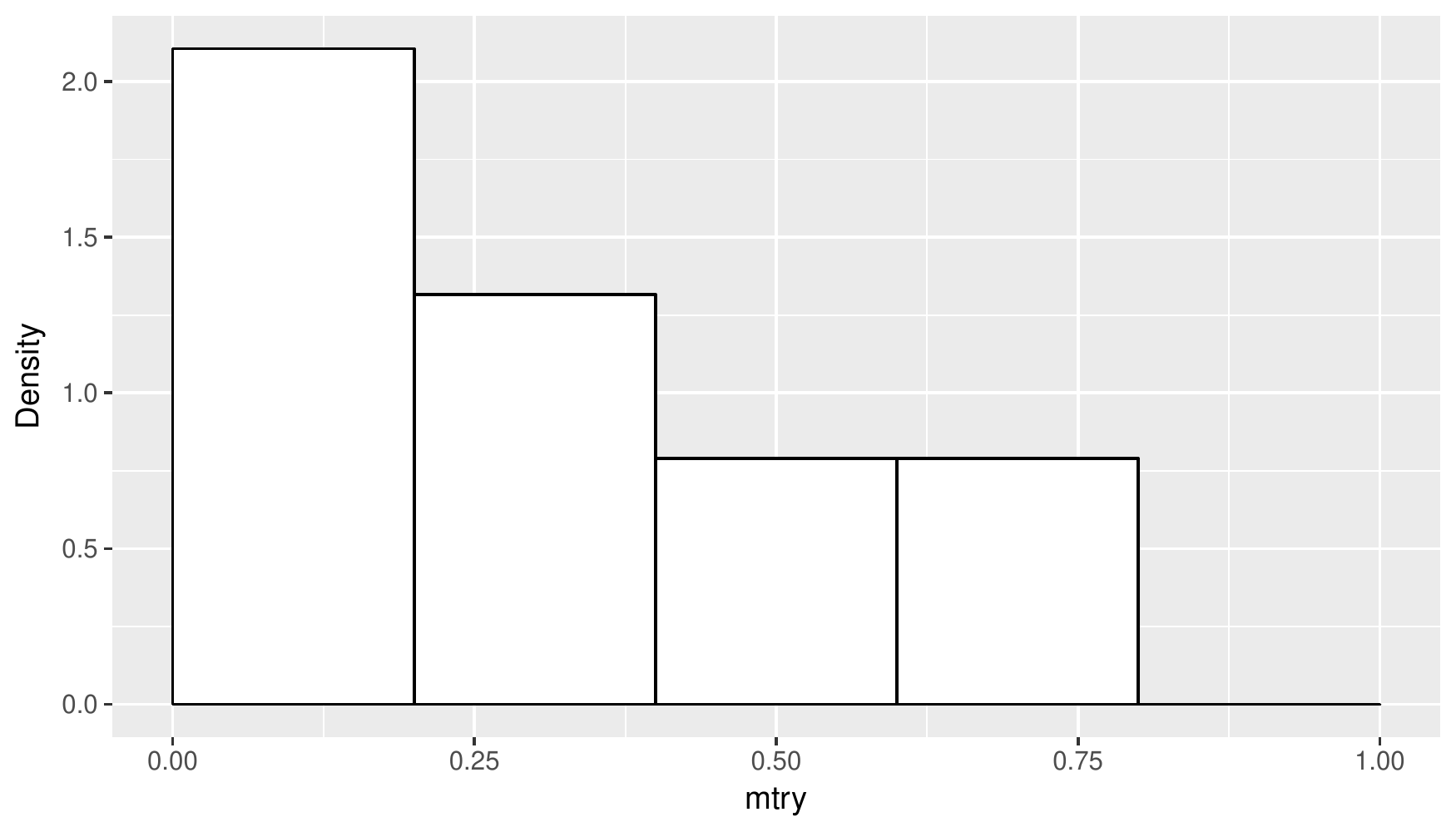} \caption[Density histogram of best parameters mtry in random forest]{Histogram of best parameter values for mtry of random forest over all considered data sets.}\label{fig:priors}
\end{figure}

As an example, Figure \ref{fig:priors} displays the full histogram of the best values of \texttt{mtry} of the random forest over all datasets. 
Note that for quite a few data sets much higher values than the package defaults seem advantageous. Analogous histograms for other parameters are available through the shiny app.

\subsubsection{Tunability of hyperparameter combinations}

As an example, Table~\ref{tab:tuna2} displays the average tunability $d_{i_1,i_2}$ of all 2-way hyperparameter combinations for \texttt{rpart}.
Obviously, the increased flexibility in tuning a 2-way combination enables larger improvements when compared with the tunability of one of the respective individual parameters. 

In Table \ref{tab:perfgain} the joint gain of tuning two hyperparameters $g_{i_1,i_2}$ instead of only the best as defined in section~\ref{sec:tunability_inter} can be seen. 
The parameters \texttt{minsplit} and \texttt{minbucket} have the biggest joint effect, which is not very surprising, as they are closely related: 
\texttt{minsplit} is the minimum number of observations that must exist in a node in order for a split to be attempted and 
\texttt{minbucket} the minimum number of observations in any terminal leaf node. 
If a higher value of \texttt{minsplit} than 
the default performs better on a dataset it is possibly not enough to set it higher without also increasing \texttt{minbucket}, so the strong relationship is quite clear. 
Again, further figures for other algorithms are available through the shiny app. 

\begin{table}[!htbp]
\centering
\begin{tabular}{lrrrr}
  \hline
 & cp & maxdepth & minbucket & minsplit \\ 
  \hline
cp & 0.002 & 0.003 & 0.006 & 0.004 \\ 
  maxdepth &  & 0.002 & 0.007 & 0.005 \\ 
  minbucket &  &  & 0.006 & 0.011 \\ 
  minsplit &  &  &  & 0.004 \\ 
   \hline
\end{tabular}
\caption{Tunability $d_{i_1,i_2}$ of hyperparameters of \texttt{rpart}, diagonal shows tunability of the single hyperparameters.} 
\label{tab:tuna2}
\end{table}

\begin{table}[!htbp]
\centering
\begin{tabular}{lrrr}
  \hline
 & maxdepth & minbucket & minsplit \\ 
  \hline
cp & 0.0007 & 0.0005 & 0.0004 \\ 
  maxdepth &  & 0.0014 & 0.0019 \\ 
  minbucket &  &  & 0.0055 \\ 
   \hline
\end{tabular}
\caption{Joint gain $g_{i_1,i_2}$ of tuning two hyperparameters instead of the most important in \texttt{rpart}.} 
\label{tab:perfgain}
\end{table}

\section{Conclusion and Discussion}
\label{sec:conclusion}

Our paper provides concise and intuitive definitions for optimal defaults of ML algorithms and the impact of tuning them either jointly, tuning individual parameters or combinations, all based on the general concept of surrogate empirical performance models.
Tunability values as defined in our framework are easily and directly interpretable as \textit{how much performance can be gained by tuning this hyperparameter?}. This allows direct comparability of the tunability values across different algorithms.

In an extensive OpenML benchmark, we computed optimal defaults for 
elastic net, decision tree, k-nearest neighbors, SVM, random forest and xgboost and quantified their tunability and the tunability of their individual parameters. 
This -- to the best of our knowledge -- has  never  been provided before in such a principled manner.
Our results are often in line with common knowledge from literature and our method itself now allows an analogous analysis for other or more complex methods. 

Our framework is based on the concept of default hyperparameter values, which can be seen both as an advantage (default values are a valuable output of the approach) and as an inconvenience (the determination of the default values is an additional analysis step and needed as a reference point for most of our measures).

We now compare our method with \citet{Rijn2017}. 
In contrast to us, they apply the functional ANOVA framework from \citet{Hutter2014} on a surrogate random forest to assess the importance of hyperparameters regarding empirical performance of a support vector machine, random forest and adaboost, which results in numerical importance scores for individual hyperparameters.
Their numerical scores are - in our opinion - less directly interpretable, but they do not rely on defaults as a reference point, which one might see as an advantage.
They also propose a method for calculating hyperparameter priors, combine it with the tuning procedure hyperband, and assess the performance of this new tuning procedure.
In contrast, we define and calculate ranges for all hyperparameters.
Setting ranges for the tuning space can be seen as a special case of a prior distribution - the uniform distribution on the specified hyperparameter space. 
Regarding the experimental setup, we compute more hyperparameter runs (around 2.5 million vs. 250000), but consider only the 38 binary classification datasets of OpenML100 while \citet{Rijn2017} use all the 100 datasets which also contain multiclass datasets. 
We evaluate the performance of different surrogate models by 10 times repeated 10-fold cross-validation to choose an appropriate model and to assure that it performs reasonably well. 

Our study has some limitations that could be addressed in the future:
a) We only considered binary classification, where we tried to include a wider variety of datasets from different domains. 
In principle this is not a restriction as our methods can easily be applied to multiclass classification, regression, survival analysis or even algorithms not from machine learning whose empirical performance is reliably measurable on a problem instance. 
b) Uniform random sampling of hyperparameters might not scale enough for very high dimensional spaces, and a smarter sequential technique might be in order here, see \citep{Bossek2015} for an potential approach of sampling across problem instances to learn optimal mappings from problem characteristics to algorithm configurations. 
c) We currently are learning static defaults, which cannot depend on dataset characteristics (like number of features, or further statistical measures) as in meta-learning. Doing so might improve performance results of optimal defaults considerably, but would require a more complicated approach. 
d) Our approach still needs initial ranges to be set, in order to run our sampling procedure. Only based on these wider ranges we can then compute more precise, closer ranges. 

\section*{Acknowledgements}
We would like to thank Joaquin Vanschoren for support regarding the OpenML platform and Andreas Müller, Jan van Rijn, Janek Thomas and Florian Pfisterer for reviewing and useful comments. Thanks to Jenny Lee for language editing.

\bibliography{tunability}

\clearpage

\section*{Appendix A. Results for accuracy and brier score}

\begin{figure}[!b]
\includegraphics[width=\maxwidth]{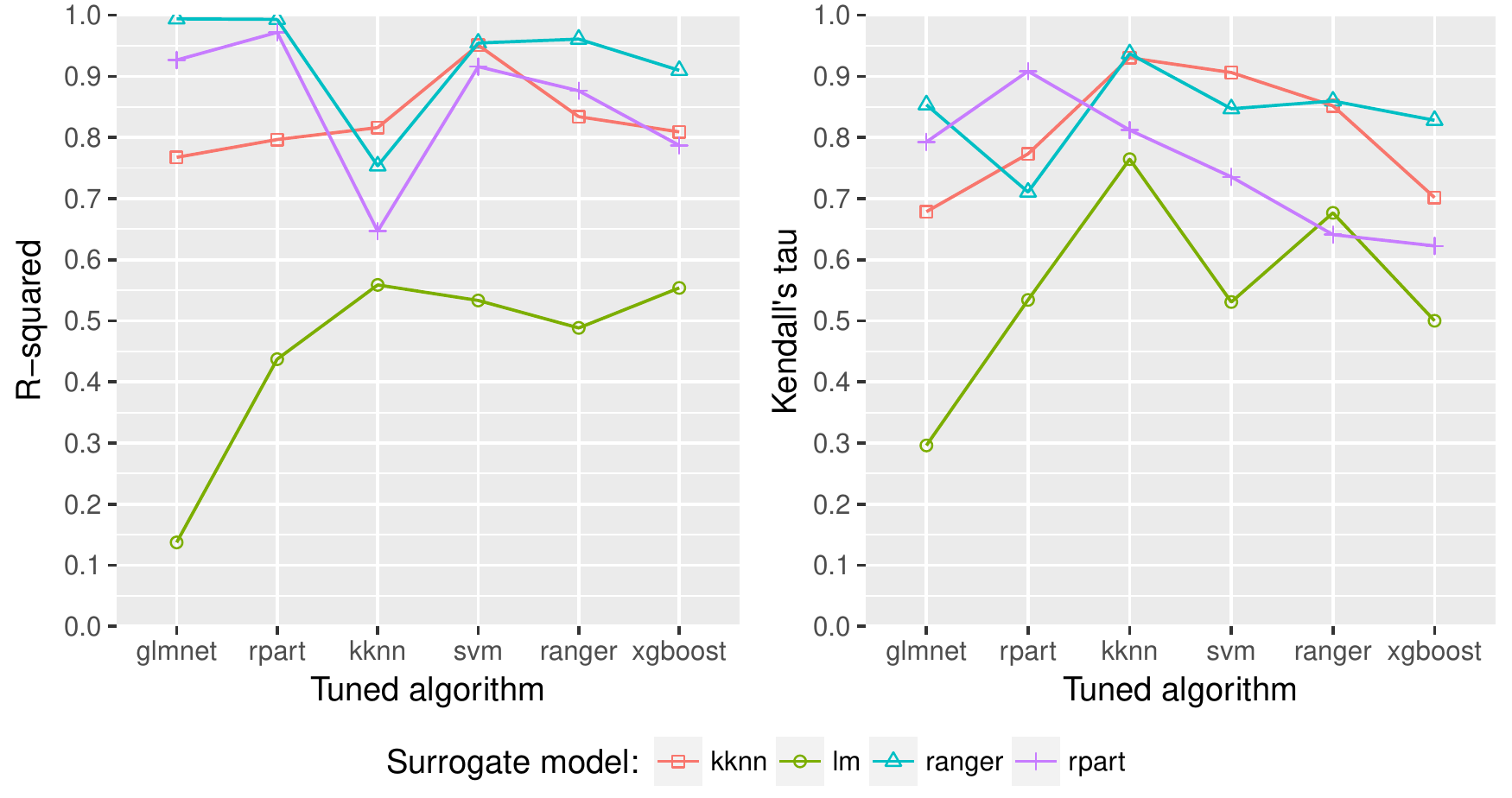} \caption[Surrogate model comparison as in figure 1 but with accuracy as target measure]{Same as figure 1 but with accuracy as target measure. Average performances over the datasets of different surrogate models (target: accuracy) for different algorithms (that were presented in \ref{sec:algorithms}).}\label{fig:surr_models_accuracy}
\end{figure}

\begin{figure}[!b]
\includegraphics[width=\maxwidth]{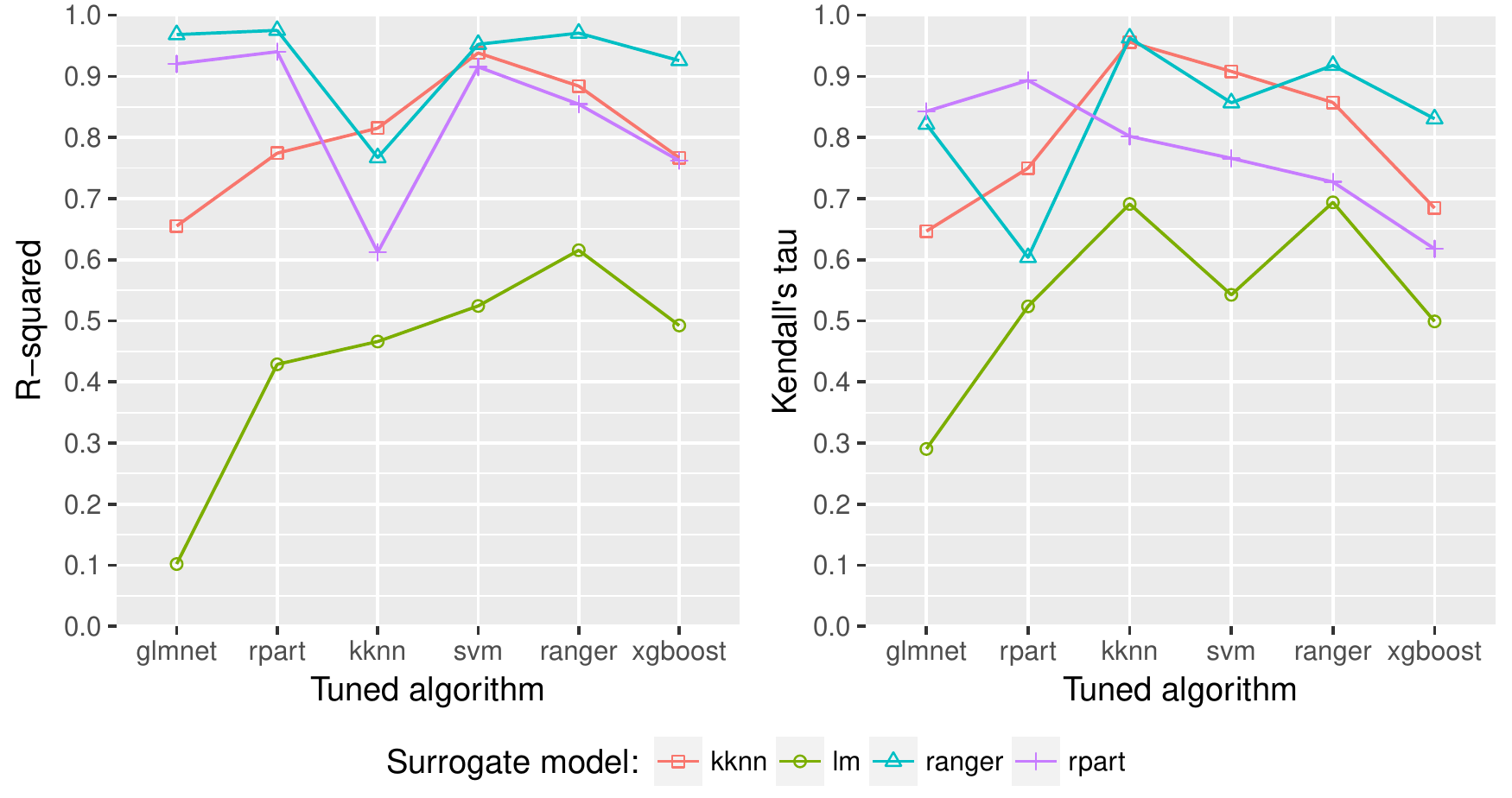} \caption[Surrogate model comparison as in figure 1 but with brier score as target measure]{Same as figure 1 but with brier score as target measure. Average performances over the datasets of different surrogate models (target: brier score) for different algorithms (that were presented in \ref{sec:algorithms}).}\label{fig:surr_models_brier}
\end{figure}

\clearpage

\begin{table}[ht]
\centering
\begin{tabular}{rrrrrr}
  \hline
Algorithm & Tun.P & Tun.O & Tun.O-CV & Improv & Impr-CV \\ 
  \hline
glmnet & 0.042 & 0.019 & 0.042 & 0.023 & 0.001 \\ 
  rpart & 0.020 & 0.012 & 0.014 & 0.008 & 0.005 \\ 
  kknn & 0.021 & 0.008 & 0.010 & 0.013 & 0.010 \\ 
  svm & 0.041 & 0.030 & 0.041 & 0.011 & -0.001 \\ 
  ranger & 0.016 & 0.007 & 0.009 & 0.009 & 0.006 \\ 
  xgboost & 0.034 & 0.011 & 0.012 & 0.023 & 0.022 \\ 
   \hline
\end{tabular}
\caption{Tunability measures as in table 2, but calculated for the accuracy. Overall tunability (regarding accuracy) with the package defaults (Def.P) and the optimal defaults (Def.O) as reference points, cross-validated tunability (Def.O-CV), average improvement (Improv) and cross-validated average improvement (Impr-CV) obtained by using new defaults compared to old defaults. The (cross-validated) improvement can be calculated by the (rounded) difference between Def.P and Def.O (Def.O-CV).} 
\label{tab:tunability_overall_acc}
\end{table}

\begin{figure}
\includegraphics[width=\maxwidth]{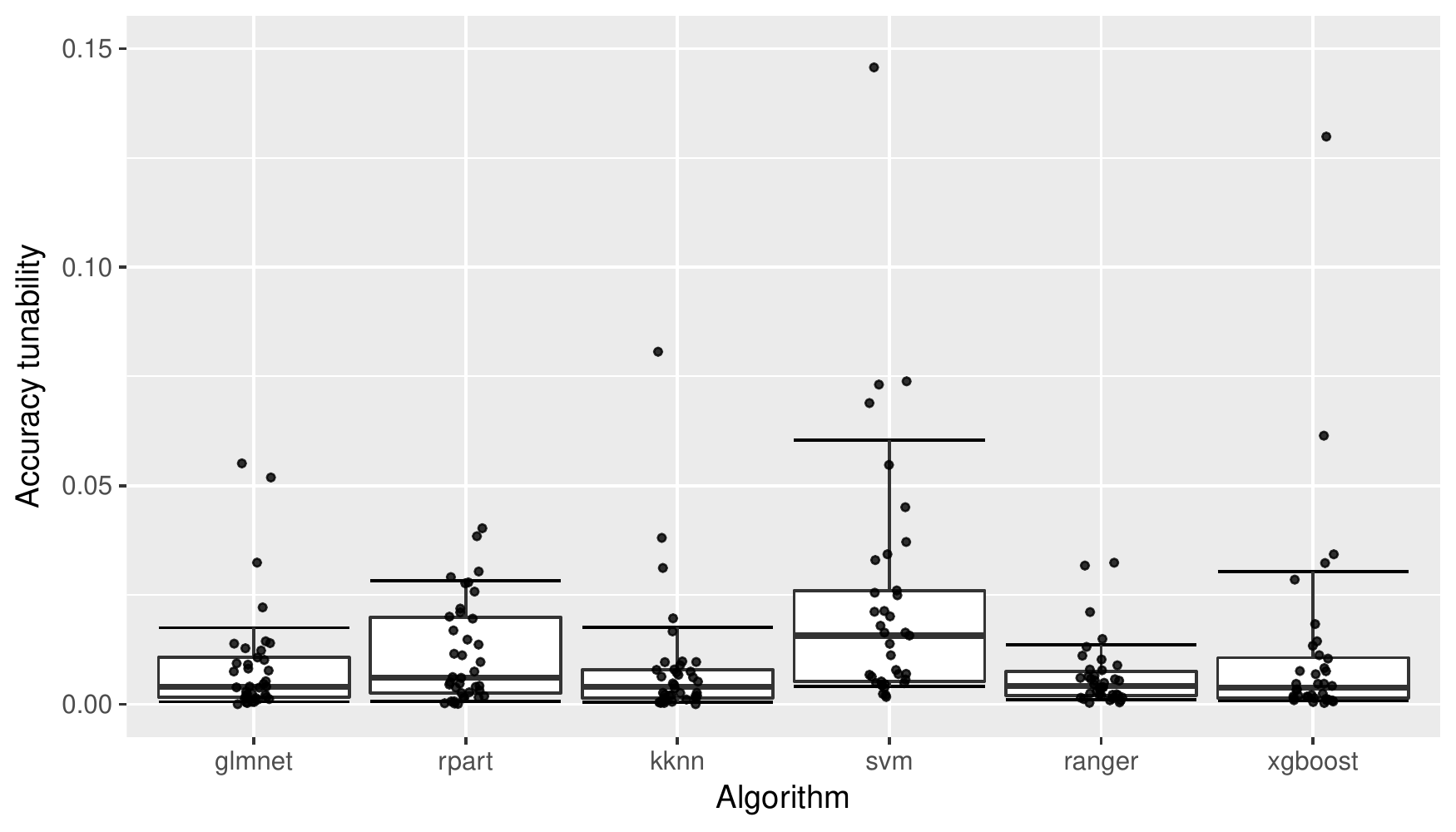} \caption{Boxplots of the tunabilities (accuracy) of the different algorithms.}\label{fig:tunability_algos_accuracy}
\end{figure}

\begin{table}
\centering
\begin{tabular}{rrrrrrr}
  \hline
Parameter & Def.P & Def.O & Tun.P & Tun.O & $q_{0.05}$ & $q_{0.95}$ \\ 
  \hline
\rowcolor[gray]{0.9}glmnet &  &  & 0.042 & 0.019 &  &  \\ 
  alpha & 1 & 0.252 & 0.022 & 0.010 & 0.015 & 0.979 \\ 
  lambda & 0 & 0.005 & 0.029 & 0.017 & 0.001 & 0.223 \\ 
  \rowcolor[gray]{0.9}rpart &  &  & 0.020 & 0.012 &  &  \\ 
  cp & 0.01 & 0.002 & 0.013 & 0.008 & 0 & 0.528 \\ 
  maxdepth & 30 & 19 & 0.004 & 0.004 & 10 & 28 \\ 
  minbucket & 7 & 5 & 0.005 & 0.006 & 1.85 & 43.15 \\ 
  minsplit & 20 & 13 & 0.002 & 0.003 & 6.7 & 47.6 \\ 
  \rowcolor[gray]{0.9}kknn &  &  & 0.021 & 0.008 &  &  \\ 
  k & 7 & 14 & 0.021 & 0.008 & 2 & 30 \\ 
  \rowcolor[gray]{0.9}svm &  &  & 0.041 & 0.030 &  &  \\ 
  kernel & radial & radial & 0.019 & 0.018 &  &  \\ 
  cost & 1 & 936.982 & 0.019 & 0.003 & 0.025 & 943.704 \\ 
  gamma & $1/p$ & 0.002 & 0.024 & 0.020 & 0.007 & 276.02 \\ 
  degree & 3 & 3 & 0.005 & 0.014 & 2 & 4 \\ 
  \rowcolor[gray]{0.9}ranger &  &  & 0.016 & 0.007 &  &  \\ 
  num.trees & 500 & 162 & 0.001 & 0.001 & 203.5 & 1908.25 \\ 
  replace & TRUE & FALSE & 0.004 & 0.001 &  &  \\ 
  sample.fraction & 1 & 0.76 & 0.003 & 0.003 & 0.257 & 0.971 \\ 
  mtry & $\sqrt{p}$ & $p\cdot0.432$ & 0.010 & 0.003 & 0.081 & 0.867 \\ 
  respect.unordered.factors & TRUE & TRUE & 0.001 & 0.000 &  &  \\ 
  min.node.size & 1 & 1 & 0.001 & 0.002 & 0.009 & 0.453 \\ 
  \rowcolor[gray]{0.9}xgboost &  &  & 0.034 & 0.011 &  &  \\ 
  nrounds & 500 & 3342 & 0.004 & 0.002 & 1360 & 4847.15 \\ 
  eta & 0.3 & 0.031 & 0.005 & 0.005 & 0.002 & 0.445 \\ 
  subsample & 1 & 0.89 & 0.003 & 0.002 & 0.555 & 0.964 \\ 
  booster & gbtree & gbtree & 0.008 & 0.005 &  &  \\ 
  max\_depth & 6 & 14 & 0.001 & 0.001 & 3 & 13 \\ 
  min\_child\_weight & 1 & 1.264 & 0.009 & 0.002 & 1.061 & 7.502 \\ 
  colsample\_bytree & 1 & 0.712 & 0.005 & 0.001 & 0.334 & 0.887 \\ 
  colsample\_bylevel & 1 & 0.827 & 0.006 & 0.001 & 0.348 & 0.857 \\ 
  lambda & 1 & 2.224 & 0.002 & 0.002 & 0.004 & 5.837 \\ 
  alpha & 1 & 0.021 & 0.003 & 0.002 & 0.003 & 2.904 \\ 
   \hline
\end{tabular}
\caption{Tunability measures for single hyperparameters and tuning spaces as in table 3, but calculated for the accuracy. Defaults (package defaults (Def.P) and own calculated defaults (Def.O)), tunability of the hyperparameters with the package defaults (Tun.P) and our new defaults (Tun.O) as reference and tuning space quantiles ($q_{0.05}$ and $q_{0.95}$) for different parameters of the algorithms.} 
\label{tab:tunability_accuracy}
\end{table}

\clearpage

\begin{table}
\centering
\begin{tabular}{rrrrrr}
  \hline
Algorithm & Tun.P & Tun.O & Tun.O-CV & Improv & Impr-CV \\ 
  \hline
glmnet & 0.022 & 0.010 & 0.020 & 0.011 & 0.001 \\ 
  rpart & 0.015 & 0.009 & 0.011 & 0.006 & 0.004 \\ 
  kknn & 0.012 & 0.003 & 0.003 & 0.009 & 0.009 \\ 
  svm & 0.026 & 0.018 & 0.023 & 0.008 & 0.003 \\ 
  ranger & 0.015 & 0.005 & 0.006 & 0.010 & 0.009 \\ 
  xgboost & 0.027 & 0.009 & 0.011 & 0.018 & 0.016 \\ 
   \hline
\end{tabular}
\caption{Tunability measures as in table 2, but calculated for the brier score. Overall tunability (regarding brier score) with the package defaults (Def.P) and the optimal defaults (Def.O) as reference points, cross-validated tunability (Def.O-CV), average improvement (Improv) and cross-validated average improvement (Impr-CV) obtained by using new defaults compared to old defaults. The (cross-validated) improvement can be calculated by the (rounded) difference between Def.P and Def.O (Def.O-CV).} 
\label{tab:tunability_overall_brier}
\end{table}

\begin{figure}
\includegraphics[width=\maxwidth]{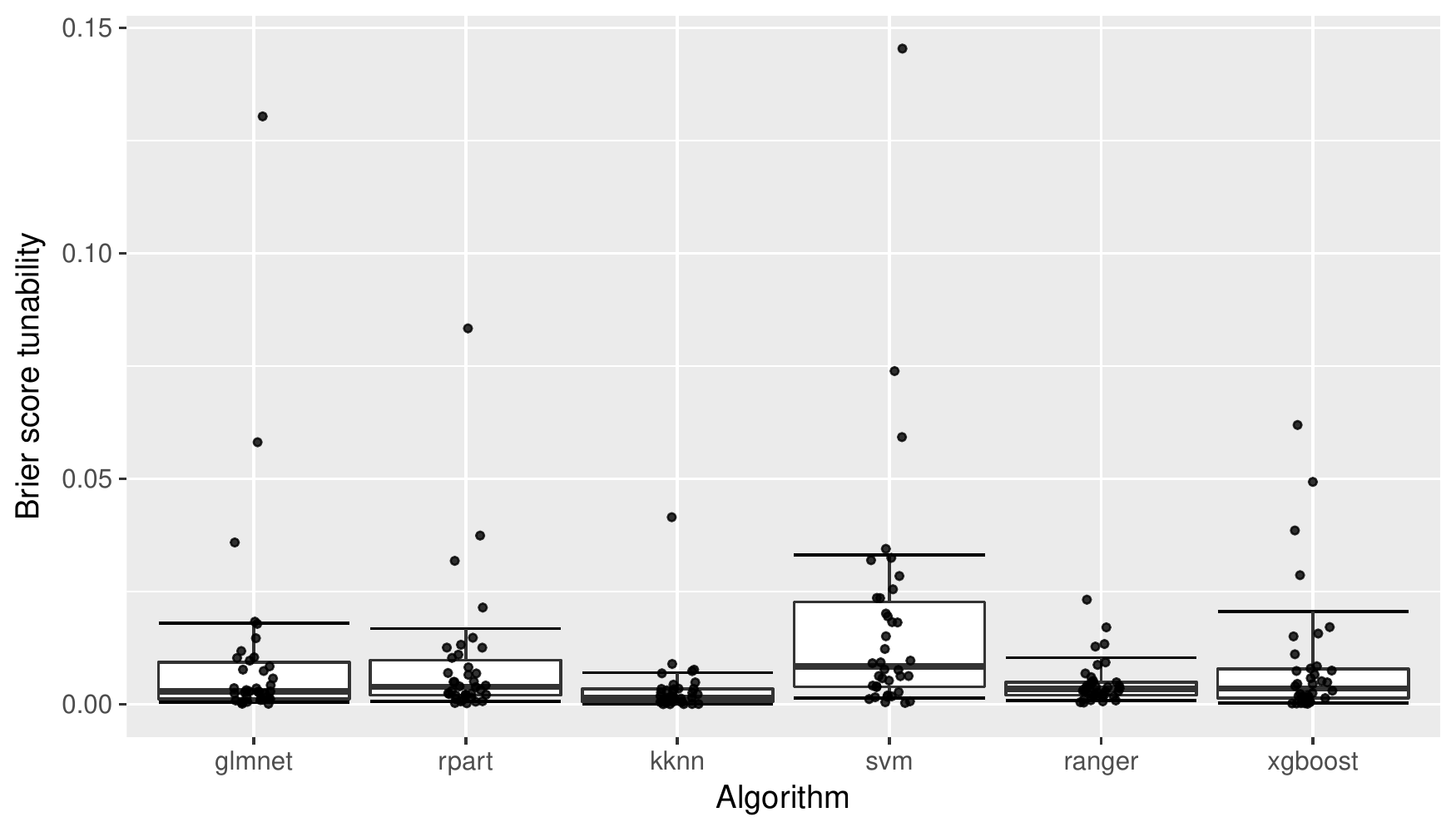} \caption[Boxplots of the tunabilities (brier score) of the different algorithms]{Boxplots of the tunabilities (brier score) of the different algorithms.}\label{fig:tunability_algos_brier}
\end{figure}

\begin{table}
\centering
\begin{tabular}{rrrrrrr}
  \hline
Parameter & Def.P & Def.O & Tun.P & Tun.O & $q_{0.05}$ & $q_{0.95}$ \\ 
  \hline
\rowcolor[gray]{0.9}glmnet &  &  & 0.022 & 0.010 &  &  \\ 
  alpha & 1 & 0.997 & 0.009 & 0.005 & 0.003 & 0.974 \\ 
  lambda & 0 & 0.004 & 0.014 & 0.007 & 0.001 & 0.051 \\ 
  \rowcolor[gray]{0.9}rpart &  &  & 0.015 & 0.009 &  &  \\ 
  cp & 0.01 & 0.001 & 0.009 & 0.003 & 0 & 0.035 \\ 
  maxdepth & 30 & 13 & 0.002 & 0.002 & 9 & 27.15 \\ 
  minbucket & 7 & 12 & 0.004 & 0.006 & 1 & 44.1 \\ 
  minsplit & 20 & 18 & 0.002 & 0.002 & 7 & 49.15 \\ 
  \rowcolor[gray]{0.9}kknn &  &  & 0.012 & 0.003 &  &  \\ 
  k & 7 & 19 & 0.012 & 0.003 & 4.85 & 30 \\ 
  \rowcolor[gray]{0.9}svm &  &  & 0.026 & 0.018 &  &  \\ 
  kernel & radial & radial & 0.013 & 0.011 &  &  \\ 
  cost & 1 & 950.787 & 0.012 & 0.002 & 0.002 & 963.81 \\ 
  gamma & $1/p$ & 0.005 & 0.015 & 0.012 & 0.001 & 4.759 \\ 
  degree & 3 & 3 & 0.003 & 0.009 & 2 & 4 \\ 
  \rowcolor[gray]{0.9}ranger &  &  & 0.015 & 0.005 &  &  \\ 
  num.trees & 500 & 198 & 0.001 & 0.001 & 187.85 & 1568.25 \\ 
  replace & TRUE & FALSE & 0.002 & 0.001 &  &  \\ 
  sample.fraction & 1 & 0.667 & 0.002 & 0.003 & 0.317 & 0.964 \\ 
  mtry & $\sqrt{p}$ & $p\cdot0.666$ & 0.010 & 0.002 & 0.072 & 0.954 \\ 
  respect.unordered.factors & TRUE & TRUE & 0.000 & 0.000 &  &  \\ 
  min.node.size & 1 & 1 & 0.001 & 0.001 & 0.008 & 0.394 \\ 
  \rowcolor[gray]{0.9}xgboost &  &  & 0.027 & 0.009 &  &  \\ 
  nrounds & 500 & 2563 & 0.004 & 0.002 & 2018.55 & 4780.05 \\ 
  eta & 0.3 & 0.052 & 0.004 & 0.005 & 0.003 & 0.436 \\ 
  subsample & 1 & 0.873 & 0.002 & 0.002 & 0.447 & 0.951 \\ 
  booster & gbtree & gbtree & 0.009 & 0.004 &  &  \\ 
  max\_depth & 6 & 11 & 0.001 & 0.001 & 2.6 & 13 \\ 
  min\_child\_weight & 1 & 1.75 & 0.007 & 0.002 & 1.277 & 5.115 \\ 
  colsample\_bytree & 1 & 0.713 & 0.004 & 0.002 & 0.354 & 0.922 \\ 
  colsample\_bylevel & 1 & 0.638 & 0.004 & 0.001 & 0.363 & 0.916 \\ 
  lambda & 1 & 0.101 & 0.002 & 0.003 & 0.006 & 28.032 \\ 
  alpha & 1 & 0.894 & 0.003 & 0.004 & 0.003 & 2.68 \\ 
   \hline
\end{tabular}
\caption{Tunability measures for single hyperparameters and tuning spaces as in table 3, but calculated for the brier score. Defaults (package defaults (Def.P) and own calculated defaults (Def.O)), tunability of the hyperparameters with the package defaults (Tun.P) and our new defaults (Tun.O) as reference and tuning space quantiles ($q_{0.05}$ and $q_{0.95}$) for different parameters of the algorithms.} 
\label{tab:tunability_brier}
\end{table}

\begin{table}
\centering
\begin{tabular}{rrrrrrr}
  Measure & \multicolumn{2}{c}{\textbf{AUC}}& \multicolumn{2}{c}{\textbf{Accuracy}} & \multicolumn{2}{c}{\textbf{Brier score}}\\ \hline
Parameter & Tun.O & Tun.O-CV & Tun.O & Tun.O-CV & Tun.O & Tun.O-CV \\ 
  \hline
\rowcolor[gray]{0.9}glmnet & 0.024 & 0.037 & 0.019 & 0.042 & 0.010 & 0.020 \\ 
  alpha & 0.006 & 0.006 & 0.010 & 0.026 & 0.005 & 0.015 \\ 
  lambda & 0.021 & 0.034 & 0.017 & 0.039 & 0.007 & 0.018 \\ 
  \rowcolor[gray]{0.9}rpart & 0.012 & 0.016 & 0.012 & 0.014 & 0.009 & 0.011 \\ 
  cp & 0.002 & 0.002 & 0.008 & 0.008 & 0.003 & 0.005 \\ 
  maxdepth & 0.002 & 0.002 & 0.004 & 0.004 & 0.002 & 0.003 \\ 
  minbucket & 0.006 & 0.009 & 0.006 & 0.007 & 0.006 & 0.006 \\ 
  minsplit & 0.004 & 0.004 & 0.003 & 0.003 & 0.002 & 0.003 \\ 
  \rowcolor[gray]{0.9}kknn & 0.006 & 0.006 & 0.008 & 0.010 & 0.003 & 0.003 \\ 
  k & 0.006 & 0.006 & 0.008 & 0.010 & 0.003 & 0.003 \\ 
  \rowcolor[gray]{0.9}svm & 0.042 & 0.048 & 0.030 & 0.041 & 0.018 & 0.023 \\ 
  kernel & 0.024 & 0.030 & 0.018 & 0.031 & 0.011 & 0.016 \\ 
  cost & 0.006 & 0.006 & 0.003 & 0.003 & 0.002 & 0.002 \\ 
  gamma & 0.022 & 0.028 & 0.020 & 0.031 & 0.012 & 0.016 \\ 
  degree & 0.014 & 0.020 & 0.014 & 0.027 & 0.009 & 0.014 \\ 
  \rowcolor[gray]{0.9}ranger & 0.006 & 0.007 & 0.007 & 0.009 & 0.005 & 0.006 \\ 
  num.trees & 0.001 & 0.002 & 0.001 & 0.003 & 0.001 & 0.001 \\ 
  replace & 0.001 & 0.002 & 0.001 & 0.002 & 0.001 & 0.001 \\ 
  sample.fraction & 0.002 & 0.002 & 0.003 & 0.003 & 0.003 & 0.003 \\ 
  mtry & 0.003 & 0.004 & 0.003 & 0.005 & 0.002 & 0.003 \\ 
  respect.unordered.factors & 0.000 & 0.000 & 0.000 & 0.001 & 0.000 & 0.000 \\ 
  min.node.size & 0.001 & 0.001 & 0.002 & 0.002 & 0.001 & 0.001 \\ 
  \rowcolor[gray]{0.9}xgboost & 0.014 & 0.017 & 0.011 & 0.012 & 0.009 & 0.011 \\ 
  nrounds & 0.002 & 0.002 & 0.002 & 0.003 & 0.002 & 0.002 \\ 
  eta & 0.005 & 0.006 & 0.005 & 0.006 & 0.005 & 0.006 \\ 
  subsample & 0.002 & 0.002 & 0.002 & 0.002 & 0.002 & 0.002 \\ 
  booster & 0.008 & 0.008 & 0.005 & 0.005 & 0.004 & 0.004 \\ 
  max\_depth & 0.001 & 0.001 & 0.001 & 0.001 & 0.001 & 0.001 \\ 
  min\_child\_weight & 0.002 & 0.003 & 0.002 & 0.002 & 0.002 & 0.003 \\ 
  colsample\_bytree & 0.001 & 0.002 & 0.001 & 0.001 & 0.002 & 0.002 \\ 
  colsample\_bylevel & 0.001 & 0.001 & 0.001 & 0.001 & 0.001 & 0.002 \\ 
  lambda & 0.002 & 0.003 & 0.002 & 0.003 & 0.003 & 0.004 \\ 
  alpha & 0.002 & 0.004 & 0.002 & 0.003 & 0.004 & 0.004 \\ 
   \hline
\end{tabular}
\caption{Tunability with calculated defaults as reference without (Tun.O) and with (Tun.O-CV) cross-validation for AUC, accuracy and brier score} 
\label{tab:tuncv}
\end{table}

\end{document}